\documentclass[final]{cvpr}
\pdfoutput=1

\usepackage{times}
\usepackage[normalem]{ulem}
\usepackage[numbers]{natbib}
\usepackage[symbol]{footmisc}
\usepackage{amsmath}
\usepackage{amssymb}
\usepackage{array,bm,multirow}
\usepackage{array}
\usepackage{booktabs}
\usepackage[labelfont=bf]{caption}
\usepackage{color}
\usepackage{graphicx}
\usepackage{enumitem}
\usepackage{mathtools}
\usepackage{microtype}
\usepackage{multirow}
\usepackage{subfigure}
\usepackage{tabularx}
\usepackage{url}
\usepackage{xcolor}
\usepackage[pagebackref=true,breaklinks=true,colorlinks,bookmarks=false]{hyperref}
\usepackage{cleveref} 
\usepackage{lipsum}

\def\cvprPaperID{****} 
\def\confYear{CVPR 2021}

\begin{document}

\title{MotionCNN: A Strong Baseline for Motion Prediction in Autonomous Driving}

\author{Stepan Konev\\
Skoltech\\
{\tt\small stevenkonev@gmail.com}
\and
Kirill Brodt\\
 Novosibirsk State University\\
{\tt\small cyrill.brodt@gmail.com}

\and
Artsiom Sanakoyeu\thanks{Now with Reality Labs at Meta.}\\
Heidelberg University\\
{\tt\small a.sanakoyeu@gmail.com}
}

\maketitle

\begin{abstract}
   To plan a safe and efficient route, an autonomous vehicle should anticipate future motions of other agents around it. Motion prediction is an extremely challenging task that recently gained significant attention within the research community. In this work, we present a simple and yet very strong baseline for multimodal motion prediction based purely on Convolutional Neural Networks. While being easy-to-implement, the proposed approach achieves competitive performance compared to the state-of-the-art methods and ranks 3rd on the 2021 Waymo Open Dataset Motion Prediction Challenge.
   Our source code is publicly available at GitHub\footnote[2]{\scalebox{1.0}{\href{https://github.com/kbrodt/waymo-motion-prediction-2021}{https://github.com/kbrodt/waymo-motion-prediction-2021}}\label{ftn:source}}.
\end{abstract}

\section{Introduction}

One of the key components of a self-driving system is motion prediction \cite{werling2010optimal,casas2018intentnet}. It is crucial for an autonomous vehicle (AV) to reliably predict future trajectories of other traffic agents, such as cars, cyclists, and pedestrians. 
However, future motion prediction and AV's route planning are still very challenging problems and are yet to be solved for an arbitrary environment scenario. 
In this paper, we tackle the motion prediction task. The most prominent approaches include image-based models which leverage birds-eye-view rasterized scene representations \cite{lee2017desire,cui2019multimodal,chai2019multipath,hong2019rules,phan2020covernet,kawasaki2021multimodal} and methods incarnated using graph neural networks \cite{casas2019spatially,gao2020vectornet,zhao2020tnt}.

We establish a simple and yet efficient motion prediction baseline based purely on Convolutional Neural Networks (CNNs). 
Our model takes a raster image centered around a target agent as input and directly predicts a set of possible trajectories along with their confidences. The raster image is obtained by rasterization of a scene and history of the all the agents. 
We evaluate our model on the 2021 Waymo Open Dataset Motion Prediction Challenge \cite{waymoOpenMotion2021} where it achieves very competitive performance: Ranks 1st using minimum average displacement error and 3rd using mAP score.
We open-source our code\footref{ftn:source} and hope that our baseline will provide a reference for future research. 


\section{Method}

\begin{figure}[t!]
\begin{center}
\includegraphics[width=1.0\linewidth]{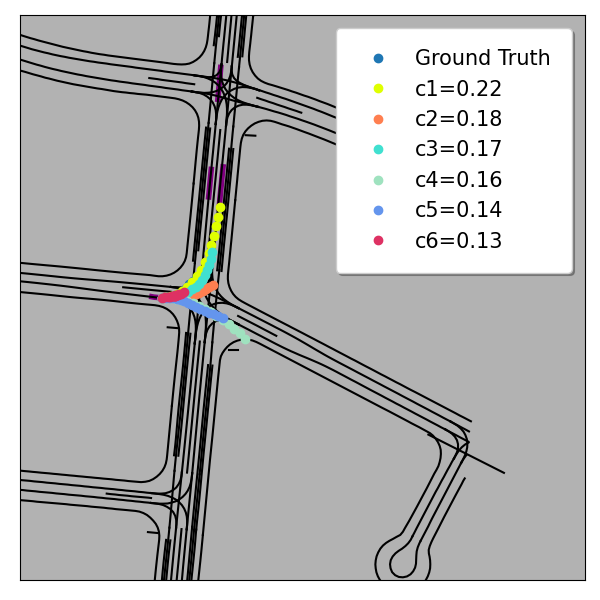}
\end{center}
   \caption{Six trajectories predicted by our model for a target agent. We visualize the trajectories using different colors and show their confidences $c_1, \dots, c_6$ in the legend.}
\label{fig:short}
\end{figure}

\begin{figure*}[t!]
\begin{center}
\includegraphics[width=1.0\linewidth]{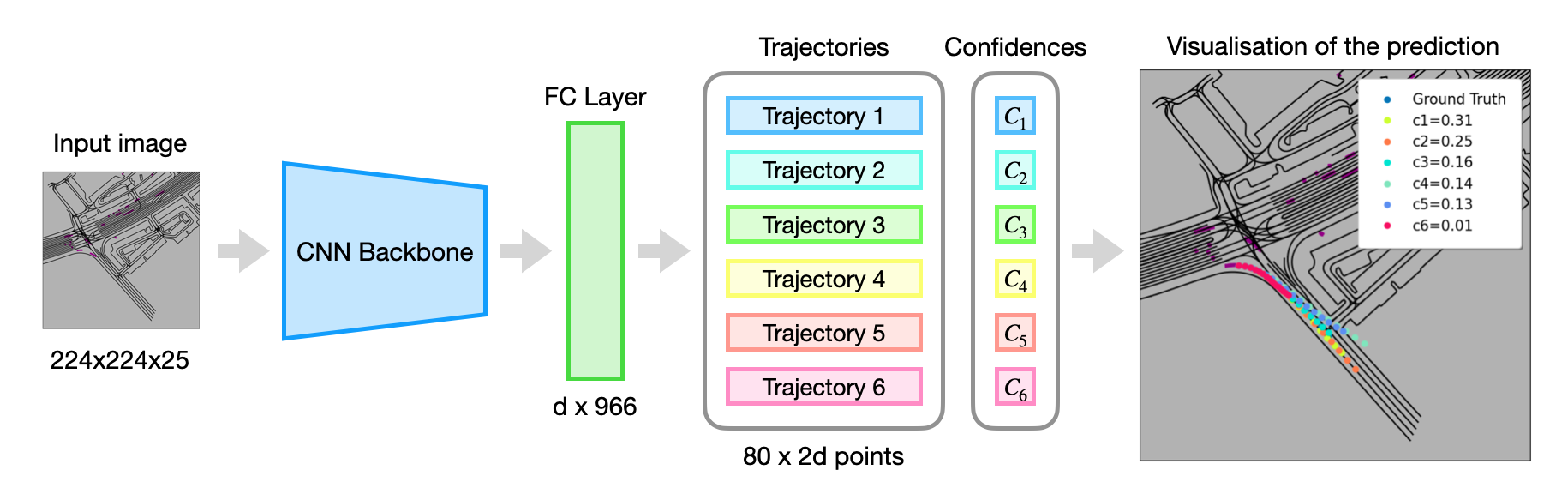}
\end{center}
   \caption{Overview of the architecture of our model.}
\label{fig:short}
\label{fig:pipeline}
\end{figure*}

We assume that object tracks are provided by some perception system \cite{qi2021offboard,yang2021auto4d} and focus only on the motion prediction. Our task is to predict the trajectory of an agent for the next $T$ seconds in the future. In this section, first, we describe how we rasterize the data and produce multi-channel images. After that we describe the architecture of our model and the loss function used for training.  

\paragraph{Rasterization}
To generate training images from raw data, we rasterize historical trajectories of the agents along with the corresponding map providing a context for the road environment. To standardise the input, we rotate and shift the frame in such a way that the target agent at the time of prediction is always located at a fixed position on the raster image and its velocity is aligned with the X-axis.

\paragraph{Model}

The future is ambiguous, so we aim to produce $K$ different hypothesis (proposals) for the future trajectory which will be evaluated against the ground truth trajectory.
We incarnate our model as an image-based regression.
Our model consists of CNN backbone pretrained on ImageNet~\cite{imagenet} with one fully-connected layer attached on top (see Fig.~\ref{fig:pipeline}). The model takes a multi-channel raster image as input and predicts $K$ trajectories along with the corresponding confidence values $c_1, \dots, c_K$, which are normalized using softmax operator such that $\sum_k c_k = 1$.


\paragraph{Loss function}
The straightforward solution would be to use a Mean Squared Error (MSE) loss. However, this loss does not allow a probabilistic modelling of multiple hypotheses and it showed poor performance in our preliminary experiments.
Instead, we propose to model possible future trajectories as the mixture of $K$ Gaussian distributions. 
In this case our network outputs the means of the Gaussians while we fix the covariance of every Gaussian in the mixture to be equal to the identity matrix $I$.

Then for the loss we can use \emph{negative log-likelihood (NLL)} of this mixture of Gaussians defined by the predicted proposals given the ground truth coordinates. In other words, given a ground truth trajectory $$X^{gt} = [(x_1, y_1), \dots ,(x_{T}, y_{T})] \\$$ and $K$ predicted trajectory hypotheses 
$$X_k = [(x_{k,1}, y_{k,1}), \dots ,(x_{k,T}, y_{k,T})],\; k=1, \dots, K,$$ we compute negative log probability of the ground truth trajectory under the predicted mixture of Gaussians with the means equal to the predicted trajectories and the identity matrix $I$ as covariance:

\begin{align*}
L &= - log P(X^{gt}) = - log \sum_k c_k \mathcal{N}(X^{gt}; \mu=X_k, \Sigma=I) \\
\end{align*}
where $\mathcal{N}(\cdot; \mu, \Sigma)$ is the probability density function for the multivariate Gaussian distribution with mean $\mu$ and covariance matrix $\Sigma$. 
The loss can be further decomposed into the product of 1-dimensional Gaussians, and we get just a logarithm of the sum of the exponents:
\begin{align*}
L &= -log \sum_k c_k \prod_{t=1}^T \mathcal{N}(x_{t}^{gt}; x_{k,t}, 1)\mathcal{N}(y_t^{gt}; y_{k,t}, 1),\\
  &= -log \sum_k e^{log(c_k) - \frac{1}{2} \sum\limits_{t=1}^T (x_t^{gt} - x_{k,t})^2 + (y_t^{gt} - y{k,t})^2}
\end{align*}
The proposed loss function does not explicitly penalize the model for producing very close trajectories. However, empirically we did not observe a mode collapse because combining all the probability mass into one mode leads to a higher risk strategy and higher loss values in case of a misprediction. Therefore, optimizing the proposed loss yields sufficient multimodality.

\paragraph{Inference}
We select the number of components in the mixture $K$ equal to the desired number of predicted hypotheses. For example, during evaluation on Waymo Open Motion Dataset~\cite{waymoOpenMotion2021} we are allowed to provide up to $6$ hypotheses of future trajectory for a target agent, so we select $K=6$. Since we model the possible space of solutions using the probability distribution, it is beneficial to produce the most diverse set of hypotheses from our distribution. One of the ways to achieve this is to simply select means of the components comprising the predicted mixture of Gaussians along with the coefficients $c_k$ as their confidences as final hypotheses for evaluation. 
While we admit that this might not be the optimal solution, we leave the exploration of other ways to sample trajectories from the predicted distribution for future work.
   
\section{Experiments}

\subsection{Dataset}
We evaluate our approach on Waymo Open Motion Dataset~\cite{waymo,waymoOpenMotion2021} by submitting our predictions to the Waymo motion prediction challenge~\cite{waymo_leaderboard_motion_21}. This dataset contains object trajectories and corresponding 3D maps for $103,354$ segments. Each segment is a $20$ seconds recording of an object trajectory at $10$Hz and map data for the area covered by the segment. A single sample comprises $1$ second of history and $8$ seconds of future data obtained by breaking the segments into 9-second windows with 5 second overlap. 
Every such sample contains up to $8$ agents marked as "valid" for which the model needs to predict their positions for $8$ seconds into the future.

\paragraph{Rasterisation details}

\begin{figure}[t]
\centering 
\includegraphics[width=1.0\linewidth]{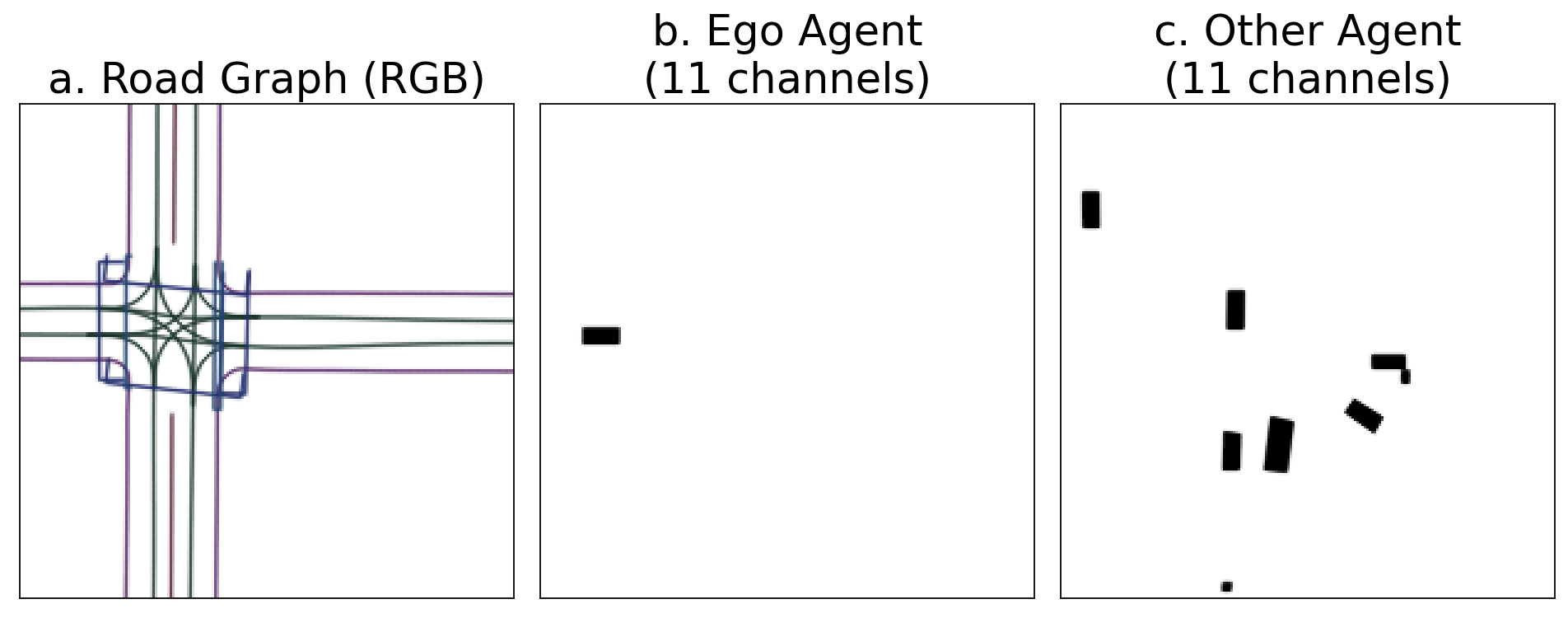}
\caption{A channel-wise illustration of the input raster of size $224\times224$ pixels with $25$ channels produced for Waymo Open Motion Dataset \cite{waymoOpenMotion2021}. The raster consists of three main blocks: (a) the first 3 channels represent the map, (b) the next $11$ channels encode the history of target object (one-channel mask per history snapshot), and, similarly, (c) another $11$ channels encode the history of all others objects.}
\label{fig:raster_structure}
\end{figure}

We create the preprocessing pipeline for Waymo Open Motion Dataset~\cite{waymoOpenMotion2021} which converts the raw data in TFRecord format to multi-channel raster images for each target object. 
For every agent we have $1$ second of history which is provided as $10$ snapshots taken at $10$Hz and a snapshot at the time of prediction (current). So, in total we have $T=11$ snapshots for every dynamic object.
We use the raster size $224 \times 224 \times (3 + 2T)$, 
where the first $3$ channels is the RGB map (road lines, crosswalks, traffic lights, etc), and every history snapshot is represented by two extra channels: (1) The mask representing the location of the target agent, and (2) the mask representing all other agents nearby (see Fig.~\ref{fig:raster_structure}).
To eliminate the redundant degrees of freedom we shift and rotate the local coordinate system in such a way that the center of the target agent is located at pixel coordinate $(61, 112)$ and its velocity is aligned with the X-axis of the image.

One of the major bottlenecks in the data pipeline is the speed of image rasterisation.
To make training faster we cache the rasterised images to disk as compressed npz files. And during training we just load them from disk instead of costly online rasterisation. This results in significant speedup enabling us to read more than a hundred images per second using a single process.

We create raster images only for the agents with the flag $state/tracks\_to\_predict$ equal to $1$ (meaning that they are "valid"). Therefore, in total, we obtain $N\approx 2.2M$ training, $192,181$ validation and $196,056$ test images.

\begin{figure}[t]
\centering
\includegraphics[width=0.4\textwidth]{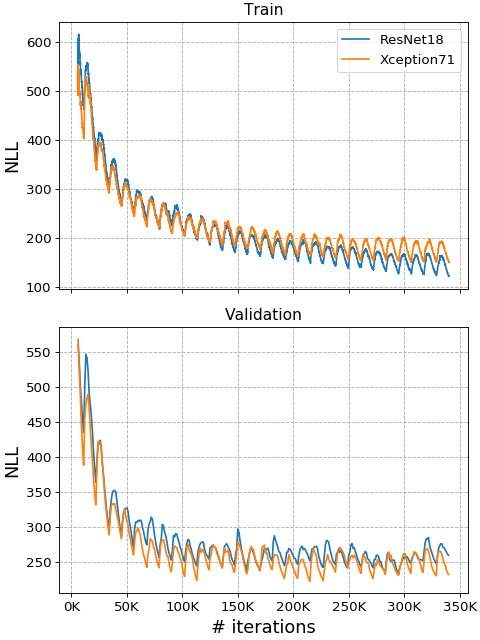}
\caption{Negative multivariate log-likelihood loss (NLL) during training iterations for MotionCNN with two different backbones.}
\label{fig:loss}
\end{figure}

\begin{table*}[t]
\vspace{-3pt}
\centering
\scalebox{0.9}{
\begin{tabular}{l|l|lllll}
& Method                & mAP             & Min ADE         & Min FDE         & Miss Rate       & Overlap Rate    \\ \toprule
\multirow{4}{*}{Test} & Waymo LSTM baseline~\cite{waymo_leaderboard_motion_21}        & 0.1756          & 1.0065          & 2.3553          & 0.3750          & 0.1898          \\ 
& ReCoAt ($2^{nd}$ place)~\cite{ReCoAt}    & 0.2711          & 0.7703          & 1.6668          & 0.2437          & 0.1642          \\
& DenseTNT ($1^{st}$ place)~\cite{denseTNT2021}  & \textbf{0.3281} & 1.0387          & 1.5514          & \textbf{0.1573} & 0.1779          \\ 
& \textbf{MotionCNN-Xception71 (Ours)} & 0.2136          & \textbf{0.7400} & \textbf{1.4936} & 0.2091          & \textbf{0.1560} \\
\midrule
\multirow{2}{*}{Val} & MotionCNN-ResNet18 (Ours) & 0.1920          & 0.8154 & 1.6396 & 0.2552          & 0.1605 \\
                     &  \textbf{MotionCNN-Xception71 (Ours)} & 0.2123 & 0.7383 & 1.4957 & 0.2072 & 0.1576 \\
\bottomrule
\end{tabular}
}
\caption{Quantitative evaluation on test and validation sets of Waymo Open Motion Dataset~\cite{waymo,waymoOpenMotion2021}.}
\label{tab:results}
\end{table*}

\begin{table*}[t]
\centering
\scalebox{0.9}{
\begin{tabular}{l|l|lllll}
&Object Type & mAP    & Min ADE & Min FDE & Miss Rate & Overlap Rate \\ \toprule
\multirow{4}{*}{Test} &Vehicle     & 0.2357 & 0.8946  & 1.8175  & 0.2138    & 0.0886       \\ 
&Pedestrian  & 0.2175 & 0.4449  & 0.9131  & 0.1276    & 0.2725       \\ 
&Cyclist     & 0.1875 & 0.8803  & 1.7501  & 0.2860    & 0.1071       \\ 
&\textbf{Avg}         & 0.2136 & 0.7400  & 1.4936  & 0.2091    & 0.1560       \\ 
\midrule
\multirow{4}{*}{Val} & Vehicle & 0.2371 & 0.8919 & 1.8154 & 0.2128 & 0.0877 \\
& Pedestrian & 0.2092 & 0.4387 & 0.9010 & 0.1254 & 0.2684\\
& Cyclist & 0.1905 & 0.8843 & 1.7707 & 0.2835 & 0.1168\\
& \textbf{Avg} & 0.2123 & 0.7383 & 1.4957 & 0.2072 & 0.1576\\
\bottomrule
\end{tabular}
}
\caption{Detailed evaluation of our MotionCNN-Xception71 model on test and validation sets of Waymo Open Motion Dataset~\cite{waymo,waymoOpenMotion2021}.}
\label{tab:results_detailed}
\end{table*}


\subsection{Metrics}
Following the evaluation protocol in~\cite{waymoOpenMotion2021}, we predict $6$ hypotheses for every target agent, but
only trajectory points subsampled at $2$Hz (which results in the subset of $16$ 2-dimensional coordinates from the predicted $80$ points) are used for computing test and validation metrics.
Average Displacement Error, Final Displacement Error (FDE) are the commonly used metrics for evaluation:
\begin{align*}
\text{ADE} = \frac{1}{T} \|X^{gt} - X \|_2, \;
\text{FDE} = \| x^{gt}_T - x_T \|_2,
\end{align*}
where $X^{gt}$ is the ground truth trajectory and $X$ is a predicted one. 
To evaluate multiple hypotheses we use minADE and minFDE:
\begin{align*}
  \text{minADE} &=\min_k\frac{1}{T} \|X^{gt} - X_k \|_2,\\ \text{minFDE} &=\min_k \|x^{gt}_T - x_{k,T} \|_2.  
\end{align*}
Additionally, following \cite{waymoOpenMotion2021}, we use a few other metrics such as Miss Rate(MR) and mean average precision (mAP). 
For the detailed explanation of these metrics we refer the reader to the work~\cite{waymoOpenMotion2021}.

\subsection{Implementation details}
Our implementation is partially based on the winning solution in Lyft Motion Prediction Challenge~\cite{kaggle_lyft_challenge2020} by Sanakoyeu et al.~\cite{sanakoyeu2021lyft}.
We use Xception71~\cite{chollet2017xception} (up to the global averaged pooling) pretrained on Imagenet as a backbone. The output of our model is $K=6$ trajectories, each containing $T=80$ 2-dimensional coordinates.
We train our model using AdamW~\cite{loshchilov2018decoupled} optimizing for $340,500$ iterations, thus using early stopping
as a regularization. We use a learning rate of ${10}^{-3}$, weight decay ${10}^{-2}$
and a batch size of $48$. We also use cosine annealing scheduler with warm restarts~\cite{loshchilov2017sgdr} every $T_0=11,350$ iterations, and with $T_{mult}=1$, $\eta_{min}={10}^{-5}$.
Training our model with Xception71~\cite{chollet2017xception} backbone took around $3$ days on a single NVIDIA V100 GPU with $32$Gb VRAM.

\subsection{Results}




Results from the final leaderboard of the Waymo open dataset motion prediction challenge~\cite{waymo_leaderboard_motion_21} are presented in Tab.~\ref{tab:results}. Despite the simplicity of the proposed approach we secured the 3rd place according to the mAP metric. Moreover, our model is superior to the other competing methods according to Min ADE, Min FDE, and Overlap Rate metrics. Note that in contrast to methods~\cite{ReCoAt,denseTNT2021}, our simple model achieves such impressive results without any use of advanced deep learning techniques or complex architectures. 

To test a more lightweight architecture, we also trained our model using ResNet18~\cite{resnet} as the backbone and evaluated it on the validation set (see Tab.~\ref{tab:results}). This architecture is 3x times faster to train than the one with Xception71 backbone, but it does not reach the same high performance showing that a sufficiently deep model is necessary for attaining good results. In Fig.~\ref{fig:loss} we show plots with train and validation loss values during training. 

In Tab.~\ref{tab:results_detailed} we also provide more detailed evaluation results for different object types separately.

\section{Conclusion}
We presented a simple yet strong baseline -- MotionCNN which is based on CNNs and produces a distribution of the hypothetical trajectories for a target agent. The proposed model is straightforward to implement and easy to train. It utilizes a birds-eye-view rasterized scene representation, which we cache as multi-channel images for faster training. We evaluated our approach on Waymo Motion Prediction Challenge~\cite{waymo_leaderboard_motion_21} where it ranked 3rd, despite being more simple then other competitors. We hope that our work will become a solid reference point for the future advancements in motion prediction.

{\small
\bibliographystyle{ieee_fullname}
\bibliography{egbib}

}

\end{document}